\documentclass[UTF8, letterpaper, 10 pt, conference]{ieeeconf}  

\IEEEoverridecommandlockouts                              
\usepackage{graphics} 
\usepackage{epsfig} 
\usepackage{mathptmx} 
\usepackage{times} 
\usepackage{amsmath} 
\usepackage{amssymb}  
\usepackage{comment} 
\usepackage{lipsum} 
\usepackage{fullpage} 
\usepackage{scrextend} 
\usepackage{listings} 
\usepackage{subcaption}
\usepackage{CJKutf8}
\usepackage{hyperref}
\usepackage{cite}
\usepackage[colorinlistoftodos]{todonotes}

\newcommand{\bW}{\mathbf{W}}

\title{\LARGE \bf
Latent Dirichlet Allocation with Residual Convolutional Neural Network Applied in Evaluating Credibility of Chinese Listed Companies
}

\author{Mohan Zhang$^{1}$, Zhichao Luo$^{2}$ and Hai Lu$^{3}$
\thanks{$^{1}$Mohan Zhang is currently a student in University of Toronto.
        {\tt\small morgan.zhang@mail.utoronto.ca}}%
\thanks{$^{2}$Zhichao Luo is currently a student in University of Toronto.
        {\tt\small }}%
\thanks{$^{3}$\href{http://www.rotman.utoronto.ca/FacultyAndResearch/Faculty/FacultyBios/Lu}{Hai Lu is currently a professor in University of Toronto.}}%
}

\begin{document}
\begin{CJK*}{UTF8}{gbsn}

\maketitle
\thispagestyle{empty}
\pagestyle{empty}


\section{Abstract}
This project demonstrated a methodology to estimating cooperate credibility with a Natural Language Processing approach. As cooperate transparency impacts both the credibility and possible future earnings of the firm, it is an important factor to be considered by banks and investors on risk assessments of listed firms.\\
This approach of estimating cooperate credibility can bypass human bias and inconsistency in the risk assessment, the use of large quantitative data and neural network models provides more accurate estimation in a more efficient manner compare to manual assessment.\\
At the beginning, the model will employs Latent Dirichlet Allocation and THU Open Chinese Lexicon from Tsinghua University to classify topics in articles which are potentially related to corporate credibility. Then with the keywords related to each topics, we trained a residual convolutional neural network with data labeled according to surveys of fund manager and accountant's opinion on corporate credibility.\\
After the training, we run the model with preprocessed news reports regarding to all of the 3065 listed companies, the model is supposed to give back companies ranking based on the level of their transparency.

\section{INTRODUCTION}

\subsection{Motivation:}
Traditionally, investigating cooperate transparency is both time consuming, inaccurate and inconsistent as the results are depending on the experience and knowledge of analysts. Therefore, we decided to develop a machine learning model where we use convolutional neural network and NLP to replace analyst in the process of transparency assessment, since a machine learning model can efficiently process a larger quantity of data compared to processing by analyst. Also, a machine learning model can ensure the result of every analysis to be consistent as long as the source of information is consistent.

\subsection{Research question and Hypothesis:}

Our research focus is the relationship between news article and corporate transparency of Chinese companies. As motivated by previous research on relationship between news articles and aspects of corporate performance, such as market performance or profit, we hypothesis that there is a connection between the corporate transparency of companies and the related news articles.

H1. There is a correlation between company description by news articles and its transparency.

On the other hand, we are expecting to verify our hypothesis through the use of Latent Dirichlet Allocation and neural network, and answer the research question:

Can we accurately evaluate a company’s corporate transparency through examining related data on news articles?

\section{Related works:}
\subsection{Apply Machine Learning into business analysis:}
Applying Artificial intelligence and machine learning techniques into business performance predicting has a long history and will continue to be an exciting research area. \textcolor{blue}{Jacky C.K. Chow et al. \cite{2018arXiv180205326C}} tried several traditional machine learning approaches and has some basic outcome. \textcolor{blue}{Jonnhyuck et al.\cite{DLLee2017}} did investigate in using Neural Network (especially CNN) to predict the performance of companies, but their approach is based on patent and some financial information. Even though above researches made significant progress trying to predict companies' performance using various of methods, they did not make full use of the most commonly accessible public information resource about trading entities. We recognized that professional news articles reporters contain information about listed companies and is able to help us analyze the credibility of listed companies. To be more specific, news reports document how listed companies interact with the market and the society, which changes public attitude towards those companies. In this way, news reports can reflect whether a listed company is trusted by the public or not, in finance, administration, social responsibility and many other ways. Although there are problems like fake news, the majority part of all news reports should be credible, thus we consider those as outliers. Therefore, the collection of all news reports (weighted by the quality) can have a high performance on predicting the credibility of listed companies. Nevertheless, it is hard for human beings to read millions of news reports generated each year to conclude whether a company is trustful. Our research hereby introduced a new methodology that use Natural Language Processing techniques with Constitutional and Residual Neural Network to process news articles, in order to predict the credibility about listed companies.\\ 
\subsection{Text summarization in Chinese articles:}
As discussed before, there are millions of news articles, which requires a method to summarize the information about listed companies' credibility. There are many text summarization models and algorithms that came into our mind when facing this task. We first tried several traditional RNN and LSTM like \textcolor{blue}{R. Nallapati et al. \cite{2016arXiv161104230N}}, then we experimented the latest non-RNN Attention way of encoding the news \textcolor{blue}{ D.  Cer et al. \cite{2018arXiv180311175C}}, which involves Transformer, presented on NIPS 2017 \textcolor{blue}{ A. Vaswani et al. \cite{2017arXiv170603762V}}. We also tested the other state-of-the-art text summarization method in the latest ACL2018 conference \textcolor{blue}{ Yen-Chun Chen et al.\cite{chen2018fast}}. However, none of them was efficient enough to deal with more than two hundred thousand articles requiring word-segmentation. It is predictable that these text summarization methods will largely limit the usability of our design due to their inefficiency on processing large corpus and mapping the summaries of each news reports to a high dimensional vector space, let alone converting them back. 
Therefore, in order to solve the Chinese word segmentation problem with large amount of data, as well as reduce non-related information, we decided to use the traditional bag of words assumption with Latent Dirichlet Allocation (LDA) proposed by \textcolor{blue}{David M. Blei et al. \cite{LDA}}. Later we discovered such traditional approach eventually yields ideal performance on both accuracy and efficiency on the text summarization of the training data. As a result, LDA was later implemented as the core component of the prepossessing of the model.

\section{Preprocessing}

\subsection{Data Source:}
Training data of the neural network in this model were made of three components, raw news Chinese articles from Jan.2017 to Sep. 2017, list of active firms from the beginning of year 2017 and financial dictionary from THU Open Chinese Lexicon.\\\\

\subsection{Word Segmentation Problem:}
The segmentation between words in Chinese phases is one of the major obstacles for processing Chinese language related contents, since Chinese words do not use space as segmentation between words. In our model, we utilized the famous bag of words assumption from Natural Language Processing, which assumes that the order of words isn't important. We also made a further assumption by observation: wording can show the potential attitude of the articles toward a target company. By assumptions above, we used the dictionary to extract important words from articles, which solved the Chinese word segmentation problem. \\
In our model, we extracted the words and their frequency in the original article only if the words are part of the THU dictionary, as the preprocessed article itself. For example, one of the article would looks like below after extraction:\\
银行/银行/银行/保险/保险/金融/市场/市场\\
the bag-of-words assumption with word-extracting using dictionary not only solves the word segmentation problem, but also get rid of "redundant" information by extracting only words contain most useful information and its frequency, as well as greatly reduce the length of the article, makes the processing much faster.\\

\subsection{Latent Dirichlet Allocation:}

Since we know that each article is characterized by a set of topics, several of those topics may be major, others may be minor. We use Latent Dirichlet Allocation to extract topics, with the corresponding topic keywords. To be more specific, we retrieve 15 topics for all the articles, each topic has 10 corresponding topic keywords. \\

\textbf{Topic keyword:} the topic keywords are the words which best describes the overall topic of articles in the training data, according to the probabilistic distribution. The order of importance is shown in descending order of the list of topics. 
for each topic, we pick the top 10 most important words that correspond with that topic. For example, if topic A corresponds with word 1,2,3,...,10 ordered by importance, then 1 is called the first keyword of topic A; 2 is called the second keyword of topic A, etc.\\

\section{Neural Network}

\textbf{Core Idea:} For each company, we summarize all news articles about that company, then represent the information as an image.\\
\subsection{Image Construction:} As we described above, we use Latent Dirichlet Allocation to preprocesses news. We divide those articles into 15 topics and pick only the 10 most important words in each topic to represent the topic. (Each topic here need not be orthogonal to each other nor have a semantical interpretation, since we only use this technique to abstract the news article, so the meaning of each topic isn't that important.) Moreover, after the preprocessing part, we have 3 sets of data for \textit{each company}: \\

\indent data1): the number of articles that mentioned that company. \\
\indent data2): the topic distribution for that company over all 15 topics. (For example, among all 1000 articles that mentioned company Z, 20 of them are most likely in topic A, 100 of them are most likely in topic B, 150 of them are most likely in topic C, etc. Then, the final topic distribution would be a vector in $\mathbb{R}^{15}$, looks like: [2\%,10\%,15\%,...]) \\
\indent data3): the number of topic keywords in all the articles that ever mentioned that company. Since there are 15 topics, each topic has 10 corresponding topic keywords, then data3 will be a 15 by 10 matrix of positive integers.\\
\\
To construct the the corresponding image for a company, we first divide data3 by data1 elementwisely, and the output will still be a 15 by 10 matrix. (We would like to do this operation because we would use data1 as a direct residual in our fully connected layer, thus we want the image we produced and the number of time a company has been referenced has zero correlation.) Next, we multiply the 15 by 10 matrix by a constant before apply the tanh function. (the constant here is a hyperparameter in order to scale integers nicely before applying tanh. The tanh function then scales everything in the range [0,1))
\begin{equation}
\mathbf{Image} = [ \mathbf(data2) \indent tanh ( constant \circ \mathbf{data3} /data1) ]
\end{equation}
Where $\circ$ stand for elementwise multiplication, / stand for elementwise division, tanh here is also performed elementwise. The final image output is a concatenation of $\mathbb{R}^{15}$ vector and 15 by 10 matrix, so the size of our image matrix is 16 by 10. Moreover, the value of each element in our image matrix is in [0,1], so we can plot the matrix as a grayscale image. Fig. 1 gives 3 example images that constructed in above way. Company a is \textit{China Life Insurance Company Limited}; company b is \textit{China Construction Bank Corporation}; company c is \textit{Shenzhen Zhongjin Lingnan Nonfemet Company Limited}. 
\begin{figure}
\caption{example images}
\centering
\begin{subfigure}[b]{0.4\textwidth}
\includegraphics[width=\textwidth]{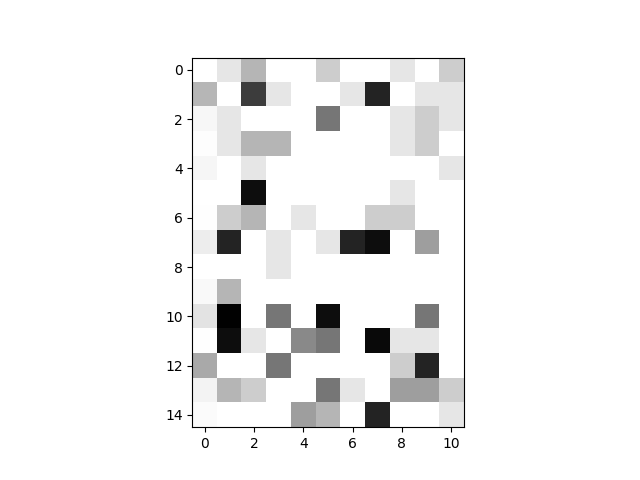}
\caption{中国人寿}
\end{subfigure}
\begin{subfigure}[b]{0.4\textwidth}
\includegraphics[width=\textwidth]{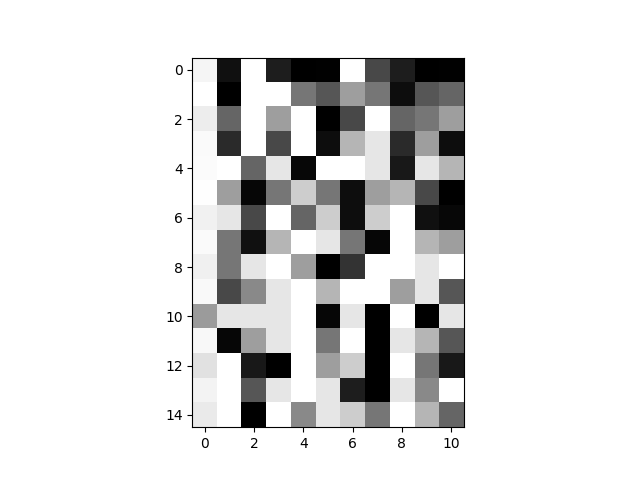}
\caption{建设银行}
\end{subfigure}
\begin{subfigure}[b]{0.4\textwidth}
\includegraphics[width=\textwidth]{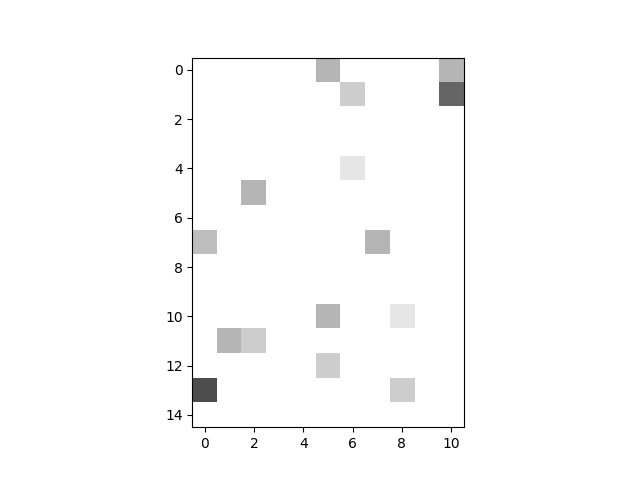}
\caption{中金岭南}
\end{subfigure}
\end{figure}\\

\subsection{Neural Network Architecture:} the build up is as follows:
\begin{equation}
\textbf{conv1} = tanh(\bW_{conv1} * \textbf{image} + \textbf{b}_{conv1})
\end{equation}
\begin{equation}
\textbf{fc1} = tanh(\bW_{fc1}^T \textbf{image} + \textbf{b}_{fc1})
\end{equation}
\begin{equation}
\textbf{fc2} = tanh(\bW_{fc2}^T \textbf{fc1} + \textbf{b}_{fc2})
\end{equation}
\begin{equation}
\textbf{fc3} = tanh(\bW_{fc3}^T \textbf{fc2} + \textbf{b}_{fc3})
\end{equation}
\begin{equation}
\textbf{fc4} = tanh(\bW_{fc4}^T [\textbf{data1} \indent \textbf{fc3}] + \textbf{b}_{fc4})
\end{equation}
\begin{equation}
\textbf{fc5} = tanh(\bW_{fc5}^T [\textbf{conv1} \indent \textbf{fc4}] + \textbf{b}_{fc5})
\end{equation}
\begin{equation}
\textbf{fc6} = \sigma(\bW_{fc6}^T \textbf{fc5} + \textbf{b}_{fc6})
\end{equation}
\begin{equation}
\textbf{output} = \sigma(\bW_{o}^T \textbf{fc6} + \textbf{b}_{o})
\end{equation}
Also, fig. 2 is the graphical architecture of the neural network.\\
 \\
\textit{Detailed Explanation:}\\
the "*" symbol stand for convolutional operation.\\
the $tanh$ symbol stand for elementwise tanh function.\\
the $\sigma$ symbol stand for elementwise sigmoid function.\\
the $[\textbf{matrix1} \indent \textbf{matrix2}]$ symbol stand for matrix concatenation.\\
\\
conv1 is the a 1-D convolutional layer, with kernel size = 10, stride = 5, aiming at capturing the mean part of topics with respect to keywords.\\
\begin{figure}
\caption{Neural Network Architecture}
\centering
\begin{subfigure}[b]{0.4\textwidth}
\includegraphics[width=\textwidth]{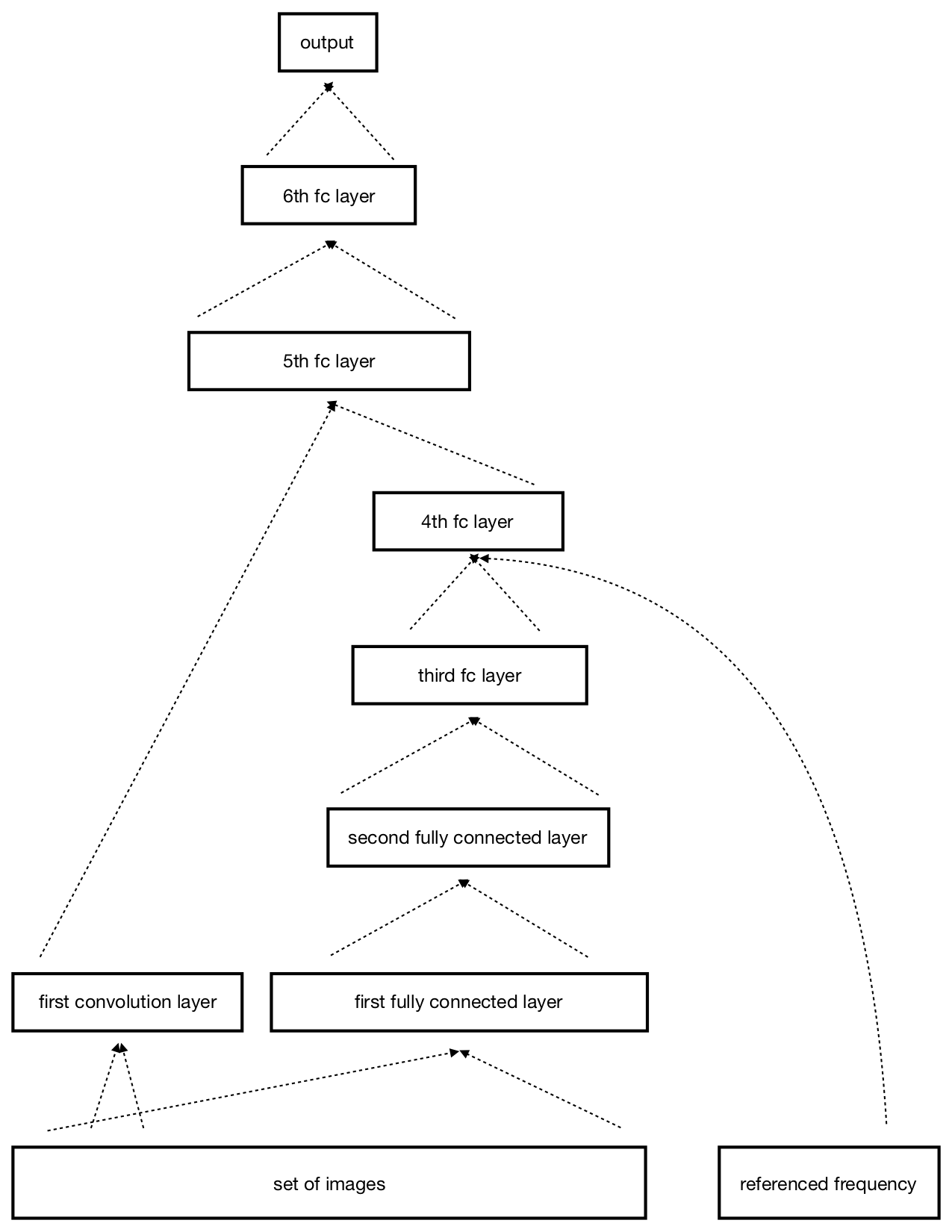}
\caption{fc stand for fully connected layer here.}
\end{subfigure}
\end{figure}\\
\subsection{Visualization of Weights:}
Now, we sum over $\bW_{fc1}$ here, ($\bW_{fc1}$ is a 16 by 10 matrix) \\
The weights are ploted in fig.3 \\
\begin{figure}
\caption{sum of weights in the first layer}
\centering
\begin{subfigure}[b]{0.4\textwidth}
\includegraphics[width=\textwidth]{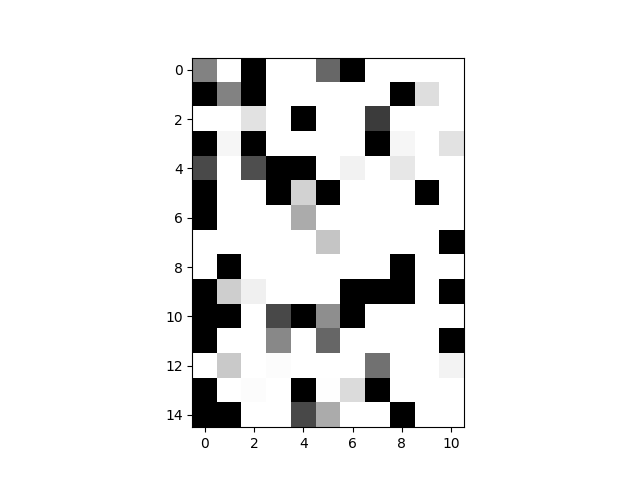}
\caption{weights for the positive prediction.}
\end{subfigure}
\begin{subfigure}[b]{0.4\textwidth}
\includegraphics[width=\textwidth]{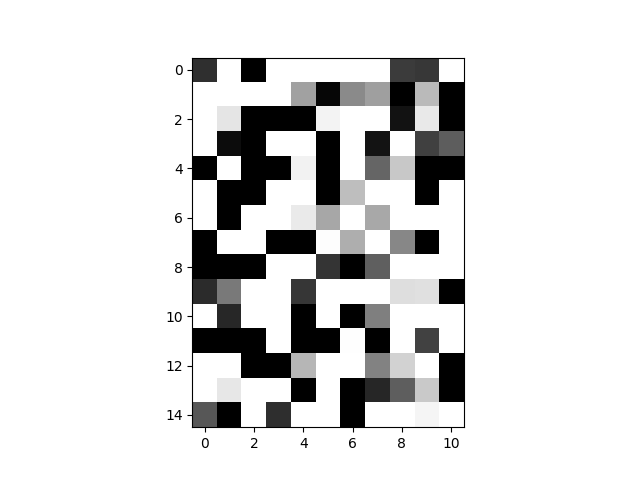}
\caption{weights for the negative prediction.}
\end{subfigure}
\end{figure}\\
\subsection{Loss function:}
for a subset of (220 out of 3065) all Chinese listed companies, we have one thousand anonymous evaluations of their credibility from professional fund managers and accountants. We calculate the metrical distance between the output of our neural network and the survey result, and regarding the metrical distance as our loss function. 
\subsection{Why we need convolutional layer?}
We hereby use convolutional layers to extract the most important words from every topic. As you may be able to see in the weight plots, the darker the pixel is in that image means the more important the word is in the context.

\subsection{Why we use residual?}
We hereby use residual, since we think the number of articles that mentioned a particular company is a key factor to the rating (it can either be a positive factor or negative one), and since we already normalized over this factor in the previous word counting part, we need to add it again. Moreover, we added it in the relatively higher level, since it is relatively more important than any single word counting and it is a more summarized information about the company.

\subsection{What is the neural network doing?}
The neural network is trying to learn from human ratings to get a general idea from the wording and word occurrence of news that, how do news articles may reflect the creditability of a company. The architecture of this neural network directly come from our research hypothsis H1.

\section{Verification:}
Verification of listed companies' credibility is generally hard, since there's actually no text-book way to measure the credibility of listed companies. Hereby we use another piece of information to build up a "negative rating system" to verify the model. As mentioned before, we use the rating of a handful of listed companies to supervise the training of our model, the higher the rating is means the more credible the company is. To verify, we also built a "negative" rating of listed companies.
Other than data we mentioned before that are used for training and test, we also have a list of companies that are investigated by the China Securities Regulatory Commission ("中国证监会"). They regularly investigate Chinese listed companies to see whether they violated any sort of regulations. The investigation procedure is not only seasonal but also depend on companies' integrity. Firstly, we have a "Date of Investigation" from the China Securities Regulatory Commission. Secondly, we have a chart of regulation-violated companies reported by China Securities Regulatory Commission. Based on these two sets of data, we train a "negative" rating of listed companies with the same model. We change the scores in our loss function from \hyperlink{Data3}{\textcolor{blue}{Data 3: Rating of a handful of listed companies}} to the investigated time of listed companies. So, hereby the higher the rating is means the less credible the company is.
The truth we already know is that there is a high correlation between companies' credibility and China Securities Regulatory Commission's investigation., Our model can be shown working if it can show this fact somehow. The verification shows that, the correlation exists between the result of our model and the negative rating. Most companies (cross-validated over 90\%) are within the rank of 200. Since there are more than three thousand companies, this is a fairly close result. So, our model is useful in the sense of predicting credibility, which answers the question rised by our research hypothesis H1.

\section{Discussion and Future Development:}
\subsection{Limitation:}
The end result of the prediction is very hard to be verified since corporate transparency is an objective opinion where varies from different organization or individuals.

\subsection{Noise and Data Credibility:}
All newspapers used as training data are treated equally regardless of the publisher’s credibility, since credibility is an objective opinion varies from different individuals.

Therefore, without taking the credibility of the publishers into account, we believe some of the articles may not contain accurate and fair articles about some companies, which may influence the accuracy of our prediction.

\subsection{Application on Other Areas:}
The methodology of our research can be applied on other investment related areas such government-issued bond, the oil market and the real estate market, since corporate transparency should also be taken into account for these kinds of investment.

Application on Credibility on Examination of Asset Management Company and Mutual Funds: Instead of examining the target of investment, the same methodology can be used to examine the credibility of mutual fund and asset management company, as investment fraud is a serious problem in China.

\bibliography{mybib}{}
\bibliographystyle{plain}


\clearpage\end{CJK*}

\end{document}